\pdfoutput=1

\documentclass[11pt, hyphens]{article}

\usepackage[]{acl}
\usepackage{times}
\usepackage{latexsym}

\usepackage[T1]{fontenc}
\usepackage{listings}
\usepackage{xcolor}

\definecolor{asparagus}{rgb}{0.53, 0.66, 0.42}
\definecolor{applegreen}{rgb}{0.55, 0.71, 0.0}
\definecolor{ao(english)}{rgb}{0.0, 0.5, 0.0}

\usepackage[utf8]{inputenc}

\usepackage{microtype}

\usepackage{graphicx}
\usepackage{svg}

\title{GRILLBot: An Assistant for Real-World Tasks with Neural Semantic Parsing and Graph-Based Representations}

\author{Carlos Gemmell, Iain Mackie, Paul Owoicho, Federico Rossetto \\
\texttt{\normalsize \{c.gemmell.1, i.mackie.1, p.owoicho.1, f.rossetto.1\}@research.gla.ac.uk}\\
\AND
Sophie Fischer, Jeffrey Dalton \\
\texttt{\normalsize \{sophie.fischer, jeff.dalton\}@glasgow.ac.uk} \\
\AND
\normalfont University of Glasgow\\
Glasgow, Scotland, UK
}

\begin{document}

\maketitle

\begin{abstract}

GRILLBot is the winning system in the 2022 Alexa Prize TaskBot Challenge, moving towards the next generation of multimodal task assistants. It is a voice assistant to guide users through complex real-world tasks in the domains of cooking and home improvement. These are long-running and complex tasks that require flexible adjustment and adaptation. The demo highlights the core aspects, including a novel Neural Decision Parser for contextualized semantic parsing, a new ``TaskGraph'' state representation that supports conditional execution, knowledge-grounded chit-chat, and automatic enrichment of tasks with images and videos. 
\end{abstract}

\section{Introduction}

We present GRILLBot, a task-oriented multimodal conversational assistant developed during the 2021/2022 Alexa Prize TaskBot Challenge \cite{own_paper}.
GRILLBot aims to be an open research platform for complex tasks and supports flexible graph-based task representations, contextual semantic parsing, and incorporates image and video content for clarity and instruction. We release the core components of the system as OAT\footnote{\url{https://github.com/grill-lab/OAT}} (Open Assistant Toolkit).

GRILLBot is still deployed throughout the United States with users able to invoke the bot by issuing the command ``Hey Alexa, Assist me'' to their voice-only or screened Alexa device. Our system provides open-ended assistance focusing in the domains of cooking and home improvement.
It guides the user through all phases of the task, from performing preference elicitation to guiding a user to a relevant task from large task corpora, (i.e. ``making a New York-style pizza" or "how to paint a wall'') and then proceeds to assist  in executing the task in an engaging way. Its capabilities include question answering, task-oriented chit-chat, and instructional video content.

GRILLBot is part of the first generation of assistants \cite{ipekalexa} that leverage screen-enabled conversational devices for complex real-world tasks. These tasks are extensive, with some taking over an hour. As a result, a performant system requires long-term state tracking with capabilities to adapt to a changing environment. It achieves this by introducing a new novel task structure, a \textit{TaskGraph}, that captures the actions and information dependencies to guide the user through a complex task. TaskGraphs are enriched offline with content from information extraction, knowledge-based content, and multimedia images and videos. 

Traditional task-oriented dialogue systems \cite{young2013pomdp} take a slot-filling approach to deriving system actions. Academic datasets such as MultiWOZ \cite{budzianowski2018multiwoz} capture slot-value pairs from the user utterances within a constrained set of domains enabling data-driven neural models. \citet{andreas2020task} extend this traditional representation towards semantic parsing with dataflow graphs while constrained to the domain of events booking in the SMCalFlow dataset. The neural decision parser in GRILLBot similarly generates code but focused on all aspects of a conversation from navigation to task search and question answering. Other challenges such as DSTC11 \cite{kottur-etal-2021-simmc} attempt this fully featured task-oriented experience, yet only do so in a virtual setting. GRILLBot stands apart as a system required to engage with real-world users in their environment and assist in complex tasks for cooking and home improvement.




\begin{figure*}[ht]
    \centering
    \includegraphics[width=\textwidth]{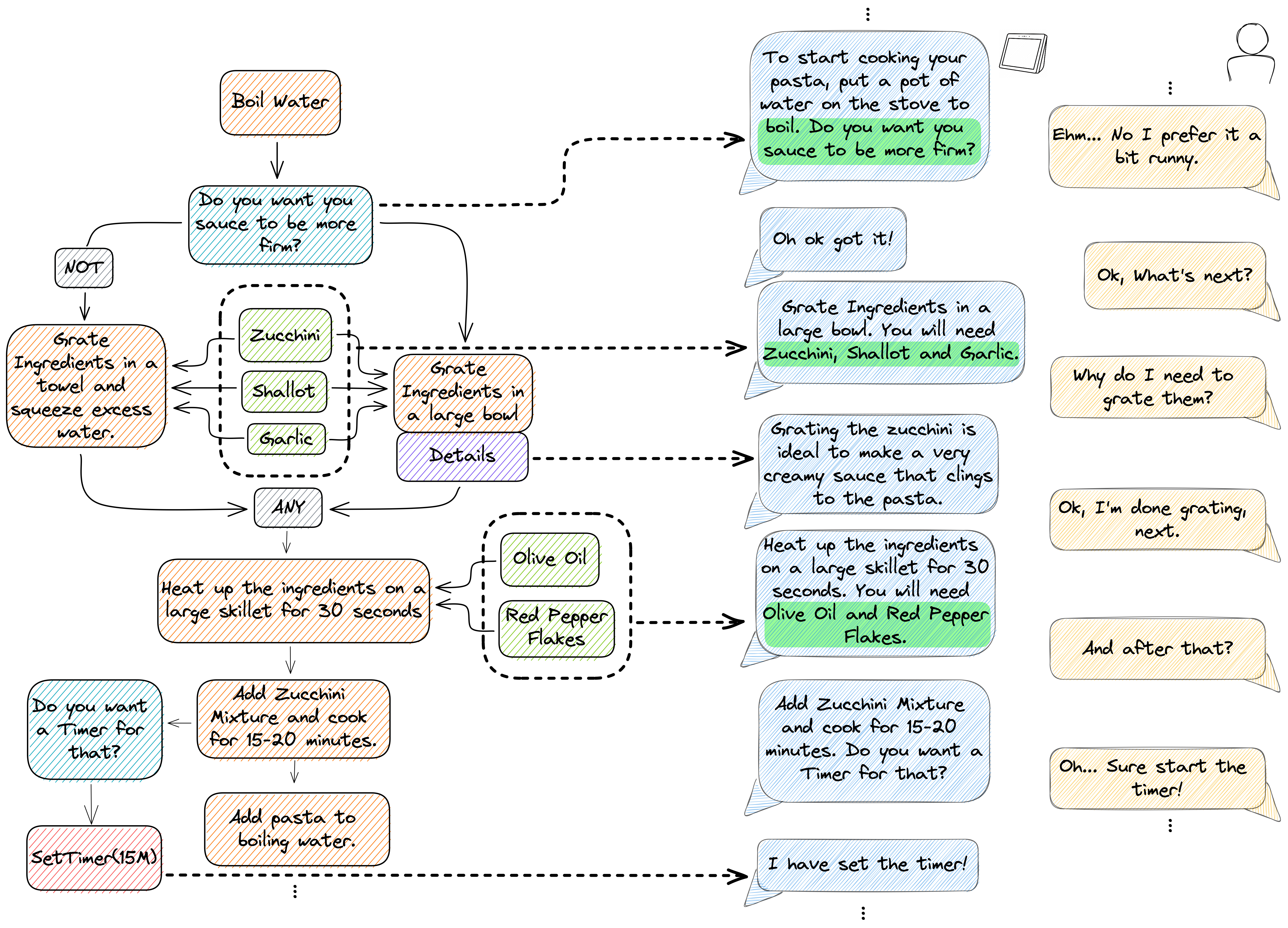}
    \caption{
    Example TaskGraph (right) and conversation (left) connect each utterance with the information in the graph nodes. 
    The figure shows how we use \textit{conditional nodes} (in light blue) to manage yes/no questions to unlock different branches of instructions.
    Conditional nodes can also unlock autonomous actions like setting a timer. 
    The figure also highlights how the \textit{requirement nodes} (in light green) are used by the system to enrich the experience by adding specific information that the user will need to perform the step. 
    Finally, the purple box highlights extra information contained inside the \textit{step nodes} to ground the QA system in domain knowledge.
    }
    \label{fig:conversation_with_graph}
    \vspace{-2mm}
\end{figure*}

\section{System overview}

The system uses a micro-service architecture with a centralized \textit{Orchestrator} that defines the system behavior. 
We use a phase-based policy to transition from searching for a suitable task (i.e. \textit{planning phase}) to guiding users in performing a task (i.e. \textit{execution phase}).

The \textit{Orchestrator} is the central process that directs and receives all the information from other microservices. 
Specifically, these child components are called \textit{functionalities} and provide the necessary tools required by policies during conversations.
The main components inside functionalities are: \textit{Neural Decision Parser}, \textit{Task Searcher} and \textit{Question Answering}.
We discuss the \textit{Neural Decision Parser} in Section \ref{sec:semantic_parser}.

The \textit{task searcher} leverages our collection of TaskGraphs to find candidate tasks. Our approach is based on a combination of traditional sparse \& dense retrieval and neural re-ranking \cite{own_paper}.
The \textit{question answering system} provides extra task information, handles user questions, and provides chit-chat elements. It uses a collection of QA systems across six categories. 





\section{TaskGraphs}

A \textit{TaskGraph} is a new graph-based representation based on a directed acyclic graph that encodes the actions and information dependencies that the system needs to enable complex dialogue flows.
Information is represented with heterogeneous nodes, each with a specific role: 
\begin{itemize}
    \item \textbf{Steps}: Represent a task instruction for the user, including visual information and textual descriptions. 
    \item \textbf{Requirements}: Represent tools and ingredients that are needed to perform the task and can be grounded to specific steps.
    \item \textbf{Conditions}: Represents yes/no gates that require external information to resolve during execution dynamically. 
    \item \textbf{Logic}: Represents logical operations. These can be used in conjunction with other nodes to enable compact dependencies and smoother execution flows. We currently support $ \land $, $\lor$ and $\neg$ operations.
    \item \textbf{Actions}: Represents actions taken by the system. This could be operations like setting a timer or adding items to a list.
    \item \textbf{Extra Information}: Represents domain/task-specific knowledge like tips or fun facts that can enrich the user experience during the execution of the task.
\end{itemize}

Combining all these nodes, we can obtain adaptive interactions where system-initiative allows the system to adapt to the user's needs. 
Figure \ref{fig:conversation_with_graph} shows how TaskGraphs can be leveraged using a task-oriented conversation helping a user cooking ``creamy zucchini pasta.''

\subsection{Offline TaskGraph Curation}

A key part of the system is providing relevant and high-quality TaskGraphs that satisfy the user's task goal. 
For example, if a user asks, ``I want to cook a gluten-free meal based around lamb shoulder``, the system must find a suitable TaskGraph.

To enable this, the system has to process rich and executable TaskGraphs offline with enough scale to cover most user needs.
Offline processing also decouples the heavy processing stages and data enrichment from online processing. 

\paragraph{Web content} 
We leverage domain experts to identify high-quality seed websites for each domain, e.g. \texttt{\normalsize wholefoodsmarket.com} and \texttt{\normalsize seriouseat.com} for cooking and \texttt{\normalsize wikihow.com} for home improvement.
We use Common Crawl to download the raw HTML for target domains and develop website-specific wrappers to extract semi-structured information about each task, i.e. title, author, description, ingredients, images, steps, ratings, videos, infoboxes, FAQs.

\paragraph{Synthesize TaskGraphs} 
The next stage of the offline process takes the semi-structured information extracted and synthesizes executable TaskGraphs.  
This creates multi-modal task nodes and connections from previously linear task steps.
For example, we can create expressive graphs that contain a summary and a detailed description for each task step, which can be accessed by users who require additional context. 
We also leverage information extraction methods, such as noun phrase detection \cite{spacy2}, to create graph connections that link required ingredients and tools to each step. 
Additionally, complex graph structure and manual augmentation can be added using a custom-developed \textit{excalidraw.com} graph interface. 
This allows loading automatically processed TaskGraphs, adding additional graph nodes and connections, and exporting the updated TaskGraphs.

\begin{figure}[ht]
    \centering
    \includegraphics[width=0.47\textwidth]{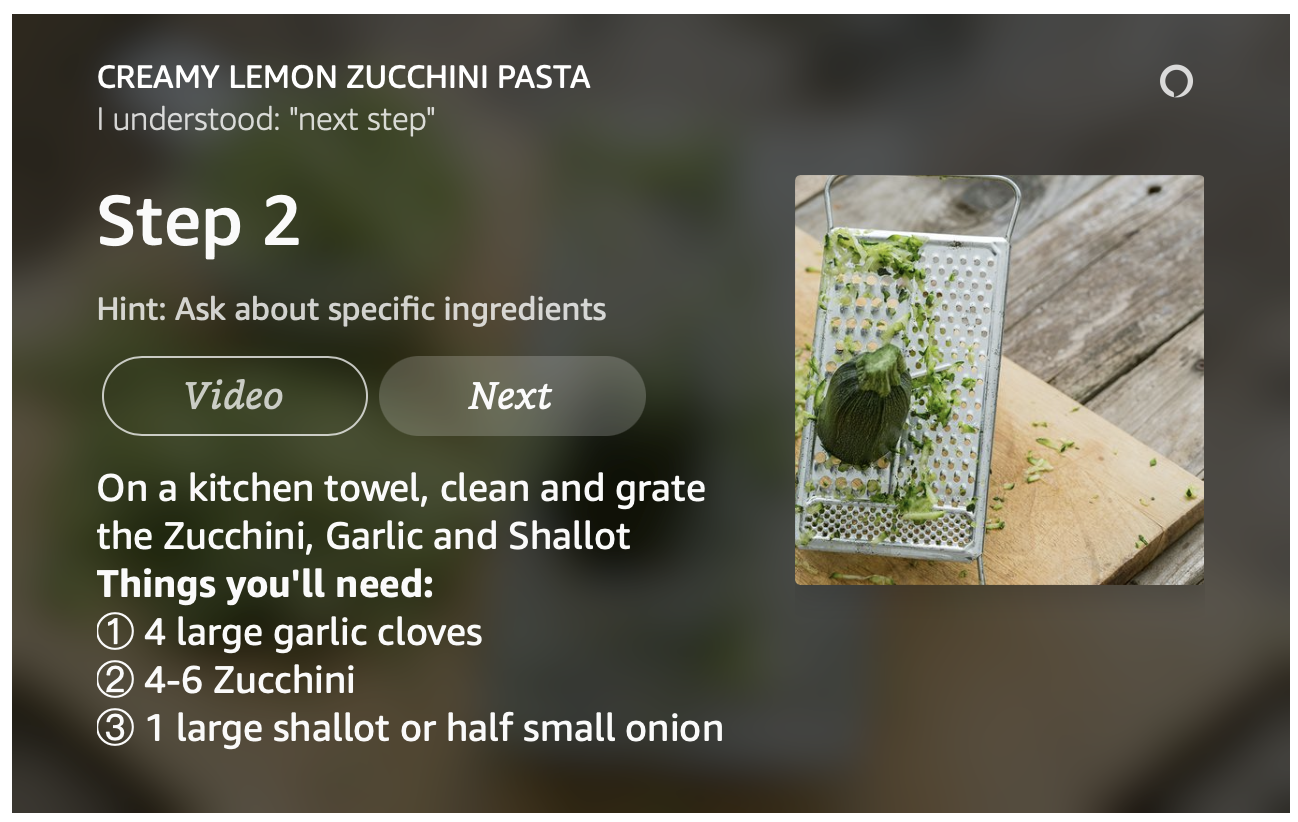}
    \caption{Multimodal UI containing text, buttons, images, and videos.}
    \label{fig:multi-modal}
\end{figure}

\paragraph{Multimodal augmentation} 
Visual information plays a crucial role in improving the success and enjoyment of users being guided through real-world tasks.
For example, showing ``How-to'' videos, images and lists of tools and ingredients offers a more compelling and useful user experience.
Figure \ref{fig:multi-modal} depicts the multi-modal experience where the screen text outlines the instruction, a list shows the ingredients required, an image enriches the user experience, and a video offers a technical demonstration.

We also develop a means of enriching task nodes if the task steps do not have aligned images and videos.
First, we extract actions (i.e "cut the beef") from a step based on a dependency parse of the step text using the spaCy toolkit.
For images, we use CLIP \cite{radford2021learning} to search over an image corpus of all other task steps images.
This uses the cosine similarity between image and step action embeddings to identify the relevant images for each step.
For videos, we develop a video corpus of domain-focused techniques, which is an index based on the video title using S-BERT \cite{reimers-2019-sentence-bert}.
Similar to image retrieval, we embed the step action as a query and rank the titles of each ``How-To'' video through a cosine similarity of the step action embedding.

\section{Neural Decision Parser}
\label{sec:semantic_parser}

\begin{figure}[ht]
    \centering
    \includegraphics[width=0.48\textwidth]{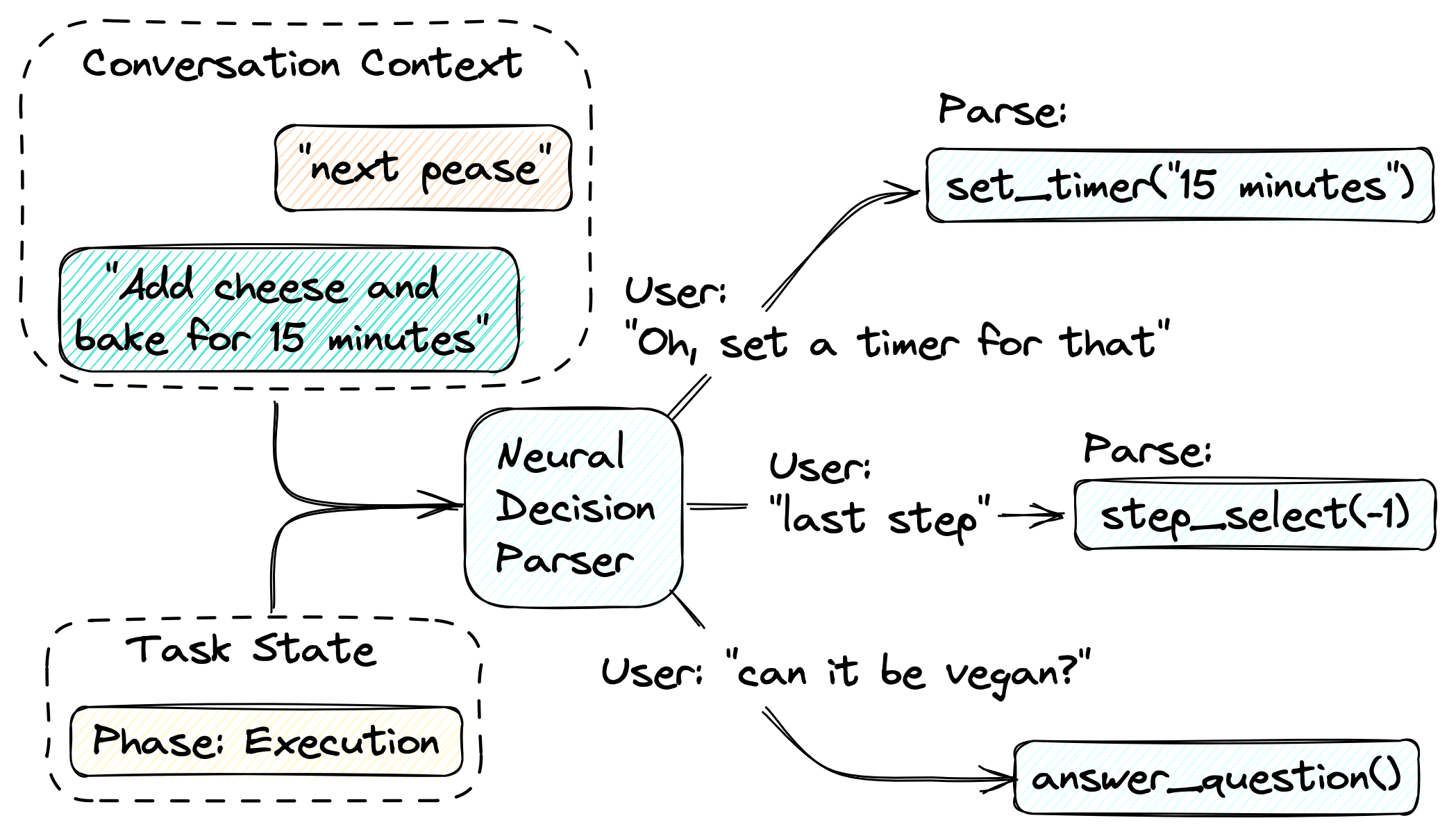}
    \caption{Example of several possible in-context parses during task execution. The Neural Decision Parser autoregressively generates the function call and arguments as code in our DSL.}
    \label{fig:conversation_with_graph}
\end{figure}

TaskGraphs allows complex representations of real-world tasks.
Due to the complex conversational dialogue required, traditional non-contextual intent classifiers struggle to manage stateful transitions.
For this reason, we develop a \textit{Neural Decision Parser} that leverages both TaskGraphs and user history for contextualized semantic parsing.   
Specifically, the model takes in natural language representations of a TaskGraph and prior conversational context to generate actions in the form of the custom GRILLBot Domain Specific Language (DSL).
These generated arguments supply the task sub-components with parsed knowledge relating to the conversation.

For example, if a user asks ``Can you go back to the first step?'', the Neural Decision Parser would generate a parameterized parse \texttt{\normalsize step\_select(1)}.

\paragraph{Contextual Semantic Parsing as a DSL}
Figure \ref{fig:dsl-example} shows our state transition domain specific language (DSL) that captures all system actions.
This DSL outlines a parameterized global command set that is understood throughout the system to derive what actions or external APIs should be called, and what the response utterance should be.
This flexible navigation allows for a complex conversation design that leverages TaskGraphs.  

\begin{figure}[ht]
\begin{lstlisting}[
    language=Python,
    commentstyle=\color{ao(english)},
    basicstyle=\small,
    backgroundcolor = \color{lightgray},
    frame=single
]
# User specifies which task to execute 
> select(option=Int) 

# Catch all for user questions
> answer_question() 

# Catch all for task search
# Vague and Theme query categories 
> search(vague=Bool, theme=String) 

# Go to prior node
> previous()

# Go to next scheduled node
> next()

# Navigate to specific task steps
> step_select(step=Int)

# Set timer with parsed time span
> timer(span=String)

# Provide details about a step
> chit_chat() 
\end{lstlisting}
\caption{A sample of the Neural Decision Parser output DSL with intent-based functions and parameterized arguments that a T5 model generates at inference time.}
\label{fig:dsl-example}
\end{figure}

\paragraph{Model}

We use a single T5 large model  \citep{raffel2020exploring}  to generate an agent action based on the TaskGraph and conversational content.
Using a pre-trained language model allows advanced language capabilities to be leveraged across all system parts, including coreference resolution, search parameterization, setting timers, and state prediction.

We train the Neural Decision Parser by annotating simulated conversations with the appropriate function calls and associated arguments. 
Our annotated training data comprises 1,200 turns across various conversation stages and includes TaskGraph and conversational context.
Through user studies and system comparisons, we find that this approach achieves strong performance, and allows flexible task navigation.

\section{Conclusion}

This demo presents GRILLBot, a newly developed Alexa Prize Taskbot system for complex real-world tasks with rich multimodal capability. 
It demonstrates multiple novel components including TaskGraphs to manage long complex tasks that are automatically enriched with offline process to add multimodal image and instructional video content. It also shows key elements of the system that make it engaging, including its flexible Neural Decision Parser that performs contextual semantic parsing as parametrized code generation.
The result is a demonstration of a new research platform designed from the ground-up around around flexible cloud micro-services and large-scale neural language models.

\bibliography{anthology,custom}
\bibliographystyle{acl_natbib}

\appendix

\end{document}